\documentclass[letterpaper, 10 pt, conference]{ieeeconf}  

\IEEEoverridecommandlockouts                              

\usepackage{caption}
\usepackage{graphicx}
\usepackage{subfigure}
\usepackage{breqn}
\usepackage{algorithm2e}
\RestyleAlgo{ruled}
\usepackage{xcolor}
\usepackage[font=small,labelfont=bf]{caption}

\title{\LARGE \bf
VLPG-Nav: Object Navigation Using Visual Language Pose Graph and Object Localization Probability Maps
}

\author{Senthil Hariharan Arul$^{\dagger*,\ddagger}$, Dhruva Kumar$^{\dagger}$, Vivek Sugirtharaj$^{\dagger}$, Richard Kim$^{\dagger}$, Xuewei (Tony) Qi$^{\dagger}$,\\ Rajasimman Madhivanan$^{\dagger}$, Arnie Sen$^{\dagger}$, Dinesh Manocha$^{\ddagger}$
\thanks{$^*$ Work done during the internship at Amazon.}%
\thanks{$^{\dagger}$ Authors are with Amazon Lab126, Sunnyvale, CA 94098, USA. {\tt\small Email: \{dhruvkm,vssugirt,richk,qixuewei,
rajasimm,senarnie\}@amazon.edu}}%
\thanks{$^{\ddagger}$ Authors are with the University of Maryland, College Park, MD 20740, USA. {\tt\small Email: \{sarul1,dmanocha\}@umd.edu}}%
}

\begin{document}

\maketitle
\thispagestyle{empty}
\pagestyle{empty}

\begin{abstract}
We present VLPG-Nav, a visual language navigation method for guiding robots to specified objects within household scenes. Unlike existing methods primarily focused on navigating the robot toward objects, our approach considers the additional challenge of centering the object within the robot's camera view. Our method builds a visual language pose graph (VLPG) that functions as a spatial map of VL embeddings. Given an open vocabulary object query, we plan a viewpoint for object navigation using the VLPG. Despite navigating to the viewpoint, real-world challenges like object occlusion, displacement, and the robot's localization error can prevent visibility. We build an object localization probability map that leverages the robot's current observations and prior VLPG. When the object isn't visible, the probability map is updated and an alternate viewpoint is computed. In addition, we propose an object-centering formulation that locally adjusts the robot's pose to center the object in the camera view. We evaluate the effectiveness of our approach through simulations and real-world experiments, evaluating its ability to successfully view and center the object within the camera field of view. VLPG-Nav demonstrates improved performance in locating the object, navigating around occlusions, and centering the object within the robot's camera view, outperforming the selected baselines in the evaluation metrics. 

\end{abstract}

\section{INTRODUCTION}

\begin{figure}[h!]
\centering     
\subfigure[Overall Approach]{\label{fig:a}\includegraphics[width=0.99\linewidth, trim={0cm, 0cm, 0cm, 0cm}, clip]{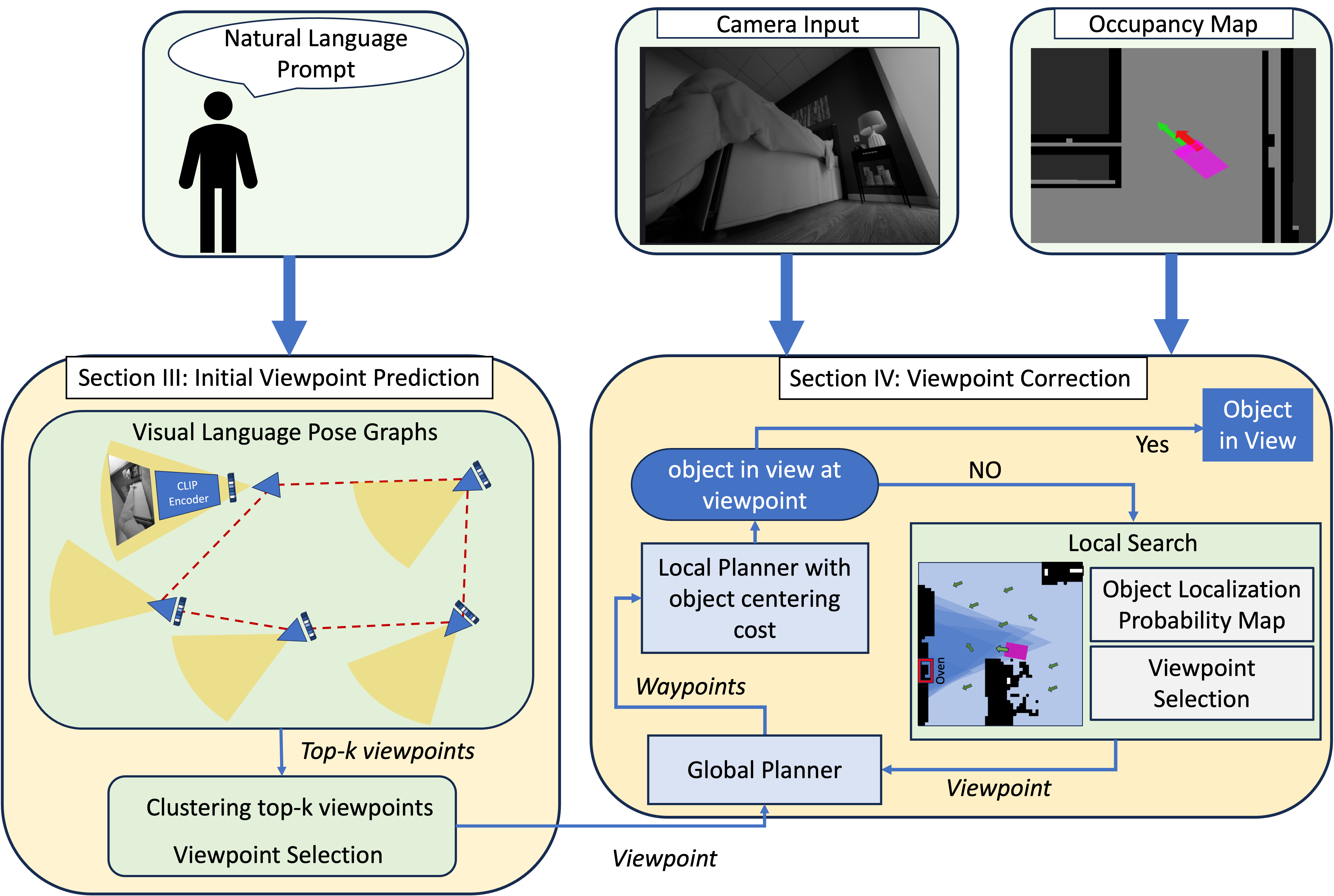}}\\
\subfigure[Initial Viewpoint]{\label{fig:a}\includegraphics[width=0.30\linewidth, trim={10cm, 0cm, 10cm, 0cm}, clip]{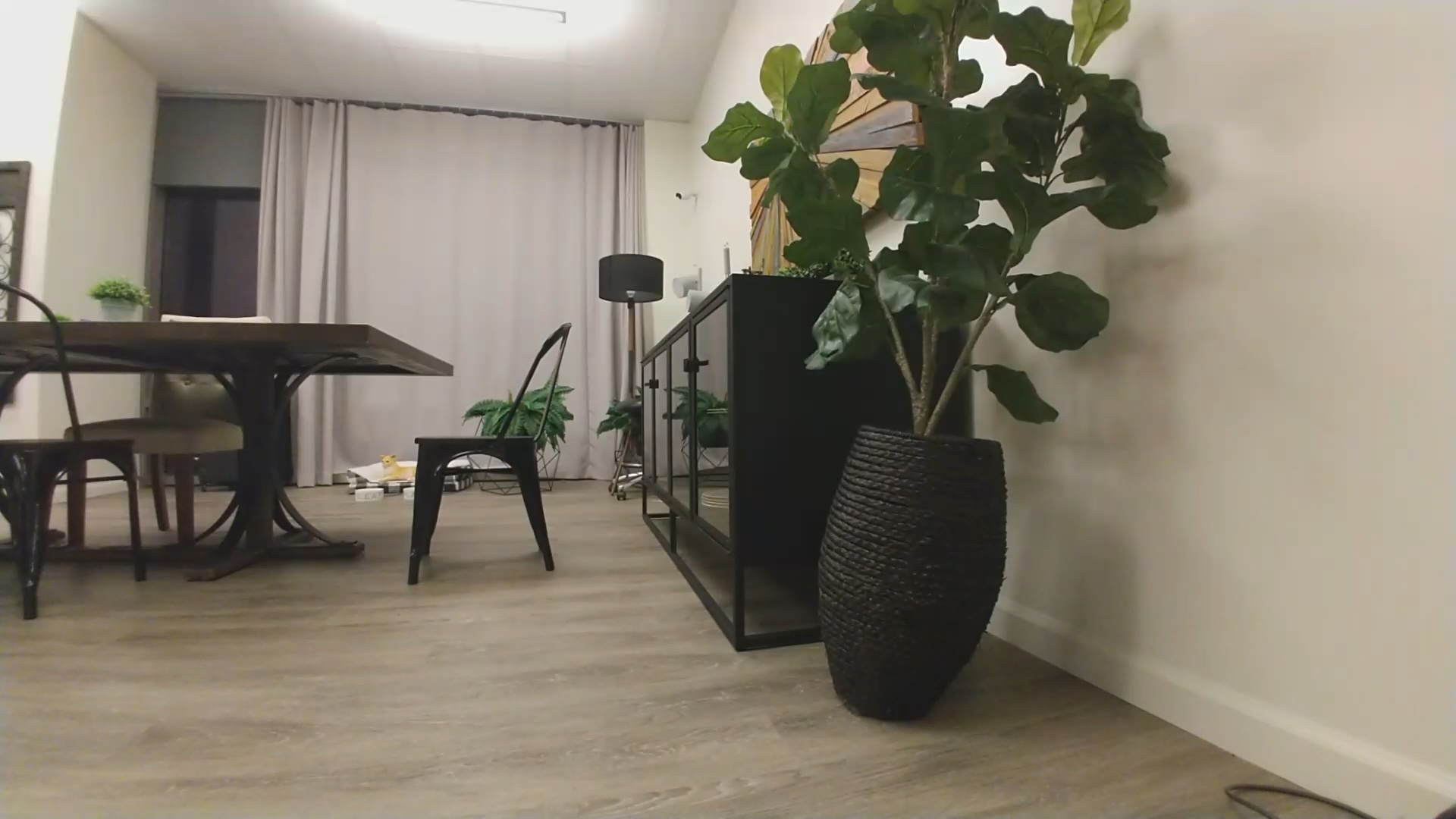}}
\subfigure[Viewpoint Occluded]{\label{fig:a}\includegraphics[width=0.32\linewidth, trim={10cm, 0cm, 7cm, 0cm}, clip]{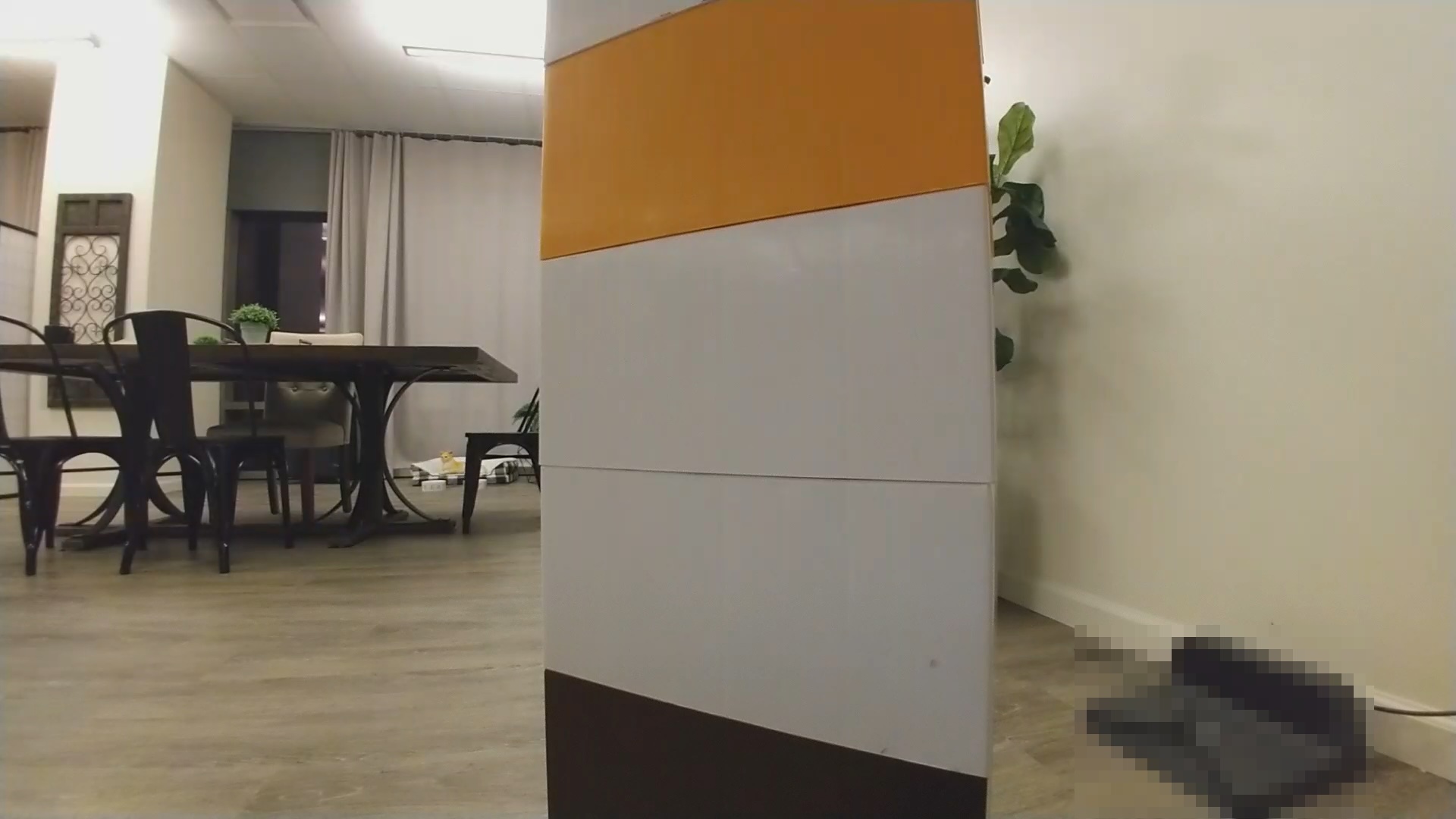}}
\subfigure[Replanned Viewpoint]{\label{fig:a}\includegraphics[width=0.34\linewidth, trim={10cm, 0cm, 5cm, 0cm}, clip]{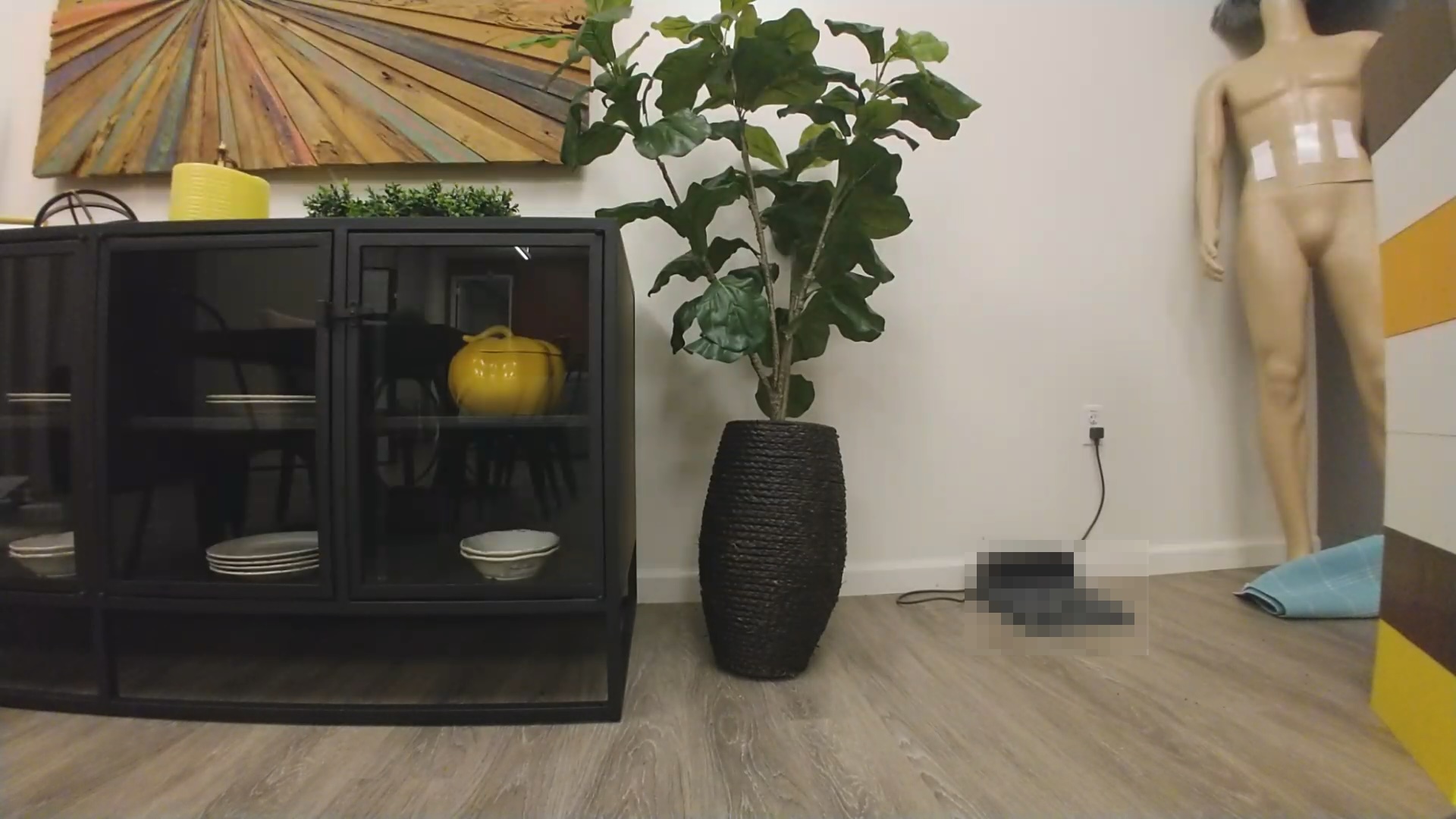}}
\caption{(Fig. 1-a) The overall framework of VLPG-Nav. Based on the object-related prompt, an initial viewpoint is computed using the visual language pose graph (VLPG) and clustering. Subsequently, we center the object in the camera image through our proposed object-centering cost function. When the object is not in view at the chosen viewpoint, we replan using an object localization probability map constructed using the local occupancy map and VLPG. (Fig. 1-b) A real-world example where the robot is tasked to view a {\em{plant}}. The figure shows the camera view from the initial viewpoint, and the robot successfully views the plant. (Fig. 1-c) In this case, we obstruct the initial viewpoint with an obstacle. The figure shows the obstructed camera view from the initial viewpoint. (Fig. 1-d) Since the object is occluded, the local search identifies an alternative viewpoint, and the robot obtains an unobstructed view of the plant.}
\vspace{-15pt}
\end{figure}

Recent advances in motion planning, perception, and deep learning have enabled robots to operate in diverse environments such as homes, factories, hospitals, and shopping malls. With their rising prevalence in day-to-day environments, there's a demand for robots to interact seamlessly with humans and undertake a broader spectrum of tasks\footnote{https://www.rd.com/article/future-robots/}. Accordingly, a crucial ability for robots is to comprehend natural language commands and leverage sensor observations to accomplish the task. 

In this regard, the problem of object navigation is especially relevant, where, given a textual prompt about an object, the robot locates and navigates to view the object. However, given the expanse of objects in a typical workspace and the ambiguity in natural language, where different words may be used for the same object, maintaining a database of all the objects (with the aliases) and their locations can be challenging. As a result, zero-shot detection methods are convenient to assist the robot in tackling such ambiguity and efficiently guide it to the object landmark of interest. 

In this paper, we consider the problem of navigating a robot to an object of interest within a household environment. Distinct from the field of object-goal navigation, we consider previously explored environments, and our focus extends beyond merely navigating the robot to the object but also to frame the object within the robot's camera view. Our focus primarily lies in household robot applications, where memory and computational efficiency are crucial. These robots often operate continuously in the same environments, allowing them to leverage prior knowledge to navigate and view the objects efficiently. 

However, relying solely on prior knowledge can be unreliable, especially considering the possibility of object occlusion or displacement. Furthermore, household environments are susceptible to changes such as the introduction of new appliances, furniture removal, or the transformation of one room into another. Additionally, localization errors and sensor noise in affordable home robots can impede the robot's ability to accurately reach predetermined viewpoints. Given these challenges, integrating current observations with prior knowledge becomes imperative to ensure task success. The method needs to be adaptable and incorporate recent information into prior knowledge, ensuring that the robot not only navigates the current environment effectively but also leverages this updated knowledge for future queries.

\subsection{Related Works}
The area of robot navigation has received significant attention over the past few decades. Classical navigation approaches~\cite{dwa,minerva,orca,dcad,park_mpepc} focus primarily on producing a safe and efficient path to a goal position. In contrast, applications such as visual question answering (VQA) require the robot to capture a specific area or object within its camera's view for task success, rather than reaching a target position goal. The area of object goal navigation focuses on this problem~\cite{ogn_1,ogn_2}, but using predefined object classes during training can fail to generalize to the spectrum of objects in a typical workspace. 

Recently, zero-shot learning methods have shown tremendous potential for computer vision applications such as image recognition~\cite{verma2018generalized,naeem2021learning}, object detection~\cite{rahman2018zero,yan2022semantics} and can be utilized to guide object goal navigation tasks. Zhao et al.~\cite{qzhao} present a zero-shot object goal visual navigation method that proposes an RL-trained semantic similarity network to prevent over-fitting to seen classes. EmbCLIP~\cite{khandelwal2022simple} leverages a contrastive language image pretraining encoder for visual navigation tasks. 

Visual-language models (VLMs)~\cite{clip,lseg,flava,glip,li2022blip,li2023blip} are networks pre-trained on internet-scale data that ground language prompts to visual observations and have shown strong generalization and reasoning ability. These models have been used for object navigation to match the robot's observation with the natural language command~\cite{lmnav,cow,vlmap,dorbala}. In LM-Nav~\cite{lmnav}, VLMs are used to match relevant landmarks in graphs based on the prompt to navigate the robot. CoW~\cite{cow} uses CLIP and GradCAM~\cite{gradcam} to detect the object of interest in the environment, but GradCAM-measured regions of interest can be noisy~\cite{vlmap}. 

VLMaps~\cite{vlmap} performs spatial grounding by generating a dense grid map of the environment with the VLM embedding as the third dimension. VLMaps have the advantage of localizing the object in 3D space accurately by using LSeg object segmentation and can be used as prior knowledge to guide future object goal navigation tasks. However, its memory requirement can be prohibitive for low-compute robots, and incorporating recent changes in the environment into the grid map may not be direct. In~\cite{esc}, they introduce common sense into a frontier-based exploration, which increases efficacy and generalizability. Du et al.~\cite{du2023object} consider object goal navigation from a navigation standpoint and propose a history-inspired navigation learning framework, and agents are observed to update states more effectively.

In many real-world applications, robots repeatedly operate in the same environment, and leveraging pre-existing knowledge can be beneficial for choosing a suitable viewpoint. 
However, these environments exhibit some level of dynamism, wherein the objects can get displaced, and the surroundings can change and potentially obstruct the view of the object. Consequently, while prior knowledge provides a good initial guess, it isn't solely reliable. To complete the object navigation task, in addition to the prior knowledge, the robot may need to locally correct and search for the object until it's in the sensor view.

\subsection{Main Contributions}

We present VLPG-Nav, an object navigation method designed for low-compute household robots. When provided with a textual prompt about an object, our approach navigates the robot to an instance of the object in the environment using prior knowledge of the environment and the robot's current observation. Given the limited computing and memory power in affordable consumer robots, we desire a memory-efficient method that allows for easy integration of new environmental information. Furthermore, the robot's final pose should provide a clear and discernible viewpoint for external observers, indicating the specific object the robot is focusing on. Our approach has three main components:

\begin{enumerate}
    \item {\bf{Initial Viewpoint Prediction:}} We construct a visual language pose graph (VLPG) during environment exploration, which is a graph of the robot's pose and the corresponding VL embedding. Given a textual prompt, we use VLPG to identify a set of relevant nodes computed based on cosine similarity, and the viewpoints are clustered to compute a suitable viewpoint guess.
    \item {\bf{Object Centering:}} To center the object in the camera image, we propose a cost function that locally corrects the robot's pose to view the object better in the image space. Specifically, the cost function orients and moves the robot toward the object of interest on reaching the initial viewpoint guess.
    \item {\bf{Local Search:}} To tackle cases when an object is undetected at the chosen viewpoint, we propose a local search method to replan an alternative viewpoint to bring the object into view. In this regard, we propose a probability map based on the VLPG clusters and the local occupancy map which provides a likelihood of the regions where the object can be localized.  
\end{enumerate}

We evaluate the effectiveness of our approach using a non-holonomic ground robot navigating toward diverse object goals in both a simulated and real-world home environment. The environment is a furnished 2-bedroom house, complete with common electronic appliances, furniture, plants, and more for a realistic test setting. We highlight the significance of the three components of our method by evaluating them in terms of pixel, and angular error in centering the object in the image space and camera field-of-view. Further, we propose a metric success weighted angular error (SAE) to compare the baselines in terms of task success.

\section{Preliminaries}
In this section, we provide an overview of various concepts used in our approach. Table~\ref{tab:symbols} summarizes the symbols and notations used in the paper.

\begin{table}[t]
    \centering
    \resizebox{0.7\linewidth}{!} {
    \begin{tabular}{c|c}
        \hline
        Symbols & Definitions \\
        \hline
        $x,y$ & 2-D position in the global frame \\
        $\theta$ & Orientation in the global frame \\
        $\mathbf{x}_r$ & Robot's Pose $(x,y,\theta)$ \\
        $\mathbf{x}_{vp}$ & A viewpoint $(x,y,\theta)$ \\
        $fov(\mathbf{x})$ & Field-of-view from pose $\mathbf{x}$ \\
        $\mathcal{C}_{vp}$ & A set of viewpoints in a cluster \\
        $\mathcal{S}_{vp}$ & A set of randomly sampled viewpoints \\
        $p_{occ}$ & Local occupancy map for the Robot\\
        \hline
    \end{tabular}}
    \caption{Symbols and their definitions}
    \label{tab:symbols}
    \vspace{-15pt}
\end{table}

\subsection{Object Navigation and Framing}
We tackle the challenge of object navigation within household environments, where the goal is for the robot to autonomously navigate and position a specified object within its camera view. Our problem differs from the related field of object-goal navigation methods in that we operate under the assumption of having prior knowledge about the environment. This is reasonable as household robots often operate repeatedly in the same environment. Furthermore, our objective extends beyond mere navigation; we aim to ensure effective capture of the object within the robot's camera view. The optimal behavior is when the object of interest is brought into the robot's camera view and centered in the image space. Given that a household environment may contain multiple instances of the same object, we consider viewing any instance of the object as a successful outcome.


\subsection{Assumption}
Natural language instructions from humans are often coarse-grained rather than a single object category. A typical query from a human is likely to be {\em{``Did I leave the stove on?"}} than {\em{``View stove."}} However, either prompt requires the robot to navigate to the same object.

In this work, we do not consider the task of breaking down the ambiguous, coarse-grained instructions (Did I leave the stove on?) into simpler keyphrases (stove). Given the recent success of Large Language Models (LLMs) (such as GPT 3~\cite{gpt3}), we assume such a model exists and provides the object-specific keyphrase as input to our approach. Moreover, at any timestep $t$, the robot can access its 2D position, orientation, camera input, and occupancy grid map of its local environment.

\subsection{Object Detection}
Grad-CAM~\cite{gradcam} a class activation mapping (CAM) approach identifies relevant regions in the image affecting the classification score. Similar to~\cite{cow}, we utilize CLIP~\cite{clip} and Grad-CAM for zero-shot object detection. We use GradCAM to create a saliency map over the input image pertaining to the input text prompt. We threshold the saliency map and construct a bounding box to localize the region of interest in the image. The bounding box information guides the local planner to center the object in the image space and to update our object localization probability map.

\subsection{Navigation Framework}
The robot navigation framework utilizes a global planner for high-level waypoint generation and a local planner for trajectory planning. Given a goal location in the environment, a global planning algorithm (Theta*~\cite{theta} in our case) generates a sequence of waypoints to the goal. The local planner, a stochastic MPC method~\cite{park_mpepc} plans the trajectories for the robot to follow the waypoints while avoiding obstacles. The local planner searches over a set of robot trajectories and selects a trajectory that minimizes a trajectory cost function. The trajectory cost function is given by:
$$
J = \sum^N p_{s_i} \cdot J_{progress} + J_{action} + (1-p_{s_i}) \cdot J_{collision}.
$$
For evaluations, a trajectory is split into $N$ segments. The progress cost measures the distance progress made over the trajectory. The action cost encourages smooth trajectories by limiting large control inputs, and the collision term limits the robot's velocity in unsafe scenarios. Here, $p_{s_i}$ is a measure of the probability that the trajectory until the $i^{th}$ segment remains collision-free. Our proposed object-centering method uses an orientation and zoom cost included in the progress term.

\subsection{Exploration}
When deployed in an unknown environment, robots lack knowledge about obstacles and the path to the goal. Exploration enables the robot to gather this information by moving around and constructing a map of the environment.
The constructed map could be referred to facilitate subsequent navigation. A popular algorithm is frontier-based exploration~\cite{frontier}, where a frontier is a portion of the region separating the explored and unexplored regions which is subsequently explored to map the environment.

\section{Initial Viewpoint Guess}

Often, robots operate continually in the same environment, allowing them to accumulate knowledge, which can be valuable information to guide object navigation. To achieve this, we employ a visual language pose graph (VLPG), a representation of the previously gathered environment knowledge from exploration, which can be leveraged during task execution to guess a robot pose highly likely to view the object of interest. This section outlines our approach for constructing VLPG and generating a guess viewpoint given an input textual prompt. 

\subsection{Visual Language Pose Graphs (VLPG)}

State-of-the-art SLAM systems use pose graphs to localize the robot and construct a spatial representation of the environment~\cite{dellaert2017factor}. The graph consists of SE(3) robot poses and images captured from the pose as nodes, while the edges are relative pose measurements. Given the comprehensive coverage of the environment during the initial mapping phase, the robot's path has transited and viewed most of the environment. This encourages using the mapping process to create prior knowledge representative of the environment. 

We build on the pose graph by augmenting each node with the VL embeddings of the robot's camera view from the pose. We call this visual language pose graph (VLPG). As the robot explores its environment, we augment the VLPG by adding a new node approximately every 100ms. 

Using VLPG for the task of viewing objects has multiple advantages:
\begin{itemize}
\item {\bf{Memory Efficiency:}} 
For an environment of $\sim 10 \times 10$ sq.m, VLPG is observed to have a graph of size $\sim 2000 \times 512$. While, a VL grid map such as in~\cite{vlmap} would produce a VL map of size $200 \times 200 \times 512$ (for a resolution of 0.05 m/cell), which is approximately 20x larger in terms of memory requirements than VLPG. Though VLPG may not localize the objects in the 3D space as accurately as in~\cite{vlmap}, we can compute a coarse localization by superimposing the camera field of view from poses that view the object (as seen in Fig.~\ref{fig:mapping}). 

\item {\bf{Easy Viewpoint Selection:}}
As VLPG nodes contain the robot's pose information, it is a collection of viewpoints, which facilitates the selection of object viewpoints. For example, given an object at a higher elevation, a suitable viewpoint is likely farther away from the object to observe it in the field of view. With the VLPG, we automatically obtain this information as it is a collection of camera viewpoints.
 
\item {\bf{Recent Environment Knowledge:}}
During long-term execution, the environment can change; new appliances may emerge, furniture can move around, etc. Hence, prior knowledge should be updated with recent information about the environment. We can directly append new nodes to the VLPG to incorporate recent information. When adding a new node, we compare the CLIP embedding of the robot's current view with the CLIP embeddings in the VLPG. The new node is added to the VLPG given it satisfies the following similarity condition, 
\begin{equation}
    \resizebox{0.9\linewidth}{!} {$
    \frac {CLIP(q*) \cdot CLIP(q)} {\left\| CLIP(q*)\right\| _{2}\left\| CLIP(q)\right\| _{2}} < \epsilon, \quad \forall q \in VLPG.$}
\end{equation}
\end{itemize}
\subsection{Generating Initial Guesses}
When deployed, the constructed VLPG is leveraged to choose a suitable viewpoint. Given an object of interest, we create two text prompts:
\begin{dmath}
   [ \texttt{A photo of \{object\}},\\
    \texttt{A photo of something else} ].
\end{dmath}
We generate a fixed-length embedding for each prompt by passing them through CLIP's text encoder and compute the cosine similarity scores between the text embedding and every camera view embedding in VLPG. The VLPG nodes resulting in a higher cosine similarity score for prompt 1 have a higher probability of viewing the object.

Though the VLPG node with the highest similarity score could be a suitable viewpoint, it does not provide information on how well it views the object. That is, the object may be only partially in view or could even be a false positive. To address this issue, we use the $top-k$ nodes with the highest similarity score. If multiple nodes view a region, we can consider the region as having high confidence in localizing the object. In this regard, we cluster the  $top-k$ nodes to remove noise (false positives) and identify clusters of poses viewing the object with a high probability.  

For every pose in the $top-k$ VL-PG nodes, we transform the viewpoint from 3-D $(x, y, \theta)$ to 4-D $(x, y, cos(\theta), sin(\theta))$ to avoid the discontinuity in orientation values. We utilize the DB-Scan algorithm to cluster the viewpoints and remove noise. The {\em{best guess}} viewpoint is chosen as the viewpoint in the largest cluster $\mathcal{C}_{vp}$ whose field of view has the maximum coverage of the region viewed by the other viewpoints in the cluster. The {\em{best guess}} is chosen as the viewpoint that maximizes the following cost function.
\begin{equation}
   {\bf{x}}_{init} =  \max_{{\bf{x}} \in \mathcal{C}_{vp}} fov({\bf{x}}) \cap \bigg( {\cup_{{\bf{y}} \in \mathcal{C}_{vp}, {\bf{y}}\neq {\bf{x}}} fov(\bf{x}}) \bigg)
\end{equation}
Here, $ {\bf{x}}_{init}$ is the chosen viewpoint. The cost function chooses a viewpoint whose field of view overlaps the most with the union of the area viewed by other viewpoints. The robot navigates towards this chosen pose to view the object.


\section{Viewpoint Correction}
In an ideal scenario, we desire the robot to reach a final viewpoint where the object of interest is clearly visible. It is equally crucial for the robot's pose to be {\em{legible}}, i.e., for an external observer, the robot's end pose must clearly indicate the chosen object of focus. Thus, as the robot navigates to the chosen viewpoint, we want the object to be in view, centered in the camera view, and occupying a significant portion of the image. However, the initial guess viewpoint may fail to satisfy this requirement due to localization uncertainty, occlusions, or object displacements. This results in two unsuccessful cases: (1) the object is in view but not centered, and (2) the object is not in view. This section outlines our approach to tackling these unsuccessful cases using current information about the robot's local environment and prior knowledge from the VLPG.

\subsection{Object in View}
As the robot nears the chosen viewpoint, the local planner uses the object location in the image to correct the robot's pose to center the object in the camera view. We use CLIP and GradCAM to localize the object in the image frame (Section 3) by constructing a bounding box for the object of interest (when detected). To perform object centering, our approach incorporates an orient and zoom cost into the local planner that aligns the object at the center and zooms in by moving closer to the object. Since the local planner optimizes over a finite time horizon, we require predictions of bounding box coordinates over the horizon. Given the bounding box center at the initial timestep, we use the image Jacobian equations~\cite{visualservo} to predict the position of the bounding box centers based on the robot's velocities over the planning horizon. 

The orient cost ($J_{orient}$) minimizes the error between the bounding box center and the image center along the horizontal image axis. 
\begin{equation}
    J_{orient} = {c_x^{t+1}}^2 - {c_x^{t}}^2
\end{equation}
Given a trajectory, the orient function is defined as the change in squared error value over the trajectory segments, with $c_x$ representing the error.

To achieve a zoomed image, the robot can move towards the object. We define the following zoom cost function.
\begin{equation}
    J_{zoom} = v \cdot \max( \cos \theta_{c}, 0) \cdot h \cdot \rho
\end{equation}
\begin{equation}
    \rho = \bigg(\frac{\max(d_{obs} - d_{thresh},0)}{d_{max}}\bigg)
\end{equation}
Here, $h$ is the timestep, and $\theta_c$ is the angle between the robot heading and the line of sight between the robot and the object center (approximated with bounding box center).

When the robot orients towards the object, the $\cos \theta_c$ component approaches one and the zoom terms become active. Until the distance to the obstacle is lower than the threshold $d_{thresh}$, the zoom cost provides a gradient to move toward the object. Ideally, we only need the zoom component if the robot is far from the object.

\subsection{Object Not-in-view}

\begin{figure}[t]
\centering
\subfigure[Object not-in-view]{\label{fig:a}\includegraphics[height=2.5cm,width=0.32\linewidth]{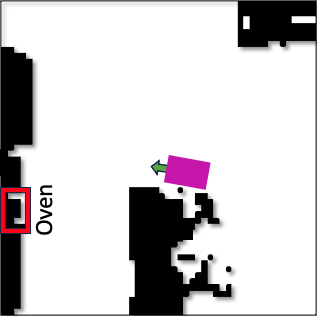}}
\subfigure[Probability Map]{\label{fig:a}\includegraphics[height=2.5cm,width=0.32\linewidth]{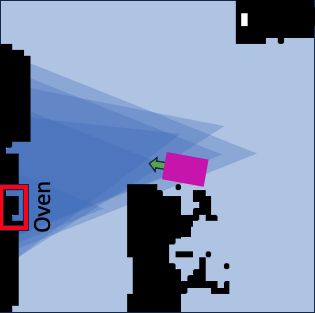}}
\subfigure[Sampled Viewpoints]{\label{fig:a}\includegraphics[height=2.5cm,width=0.32\linewidth]{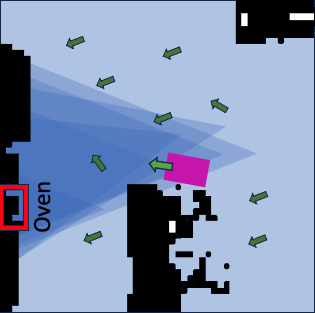}}
\caption{{\bf{Local Search: }} We illustrate a scenario with the robot tasked with viewing an oven. (Fig. 3-a) The initial viewpoint guess results in the robot's view of the oven being occluded. (Fig. 3-b) Based on the viewpoint cluster obtained from VLPG and clustering, we compute a probability map for localizing the oven. The obstacles are represented in black; regions with higher intensity blue are more likely to localize the object. (Fig. 3-c) Using the probability map, we sample a set of viewpoints from which a suitable alternative viewpoint is chosen based on our optimization objective.}
\vspace{-5pt}
\end{figure}

In certain cases, the initial guess viewpoint may fail to bring the object into the camera's view, and the robot may need to search the surrounding environment to locate it. In this work, we assume the region around the initial viewpoint to be more likely to localize the object of interest, and we utilize the viewpoint clusters from VLPG to construct an object localization probability map ($p_l$). Consequently, we use the probability map to compute alternative viewpoints to view the higher probability regions. 

First, we initialize a probability map $p_{nl}$, which is the probability of not localizing the object. The probability map is identical in size to the robot's local occupancy map ($p_{occ}$), a $6m \times 6m$ grid map, with its grid cell initialized as
$$
  p_{nl}(i,j) = {
  \begin{cases}
    1, & \text{for } p_{occ}(i,j) = 1 \\
    0.9, & \text{for } else 
  \end{cases}}.
$$
Here, $i,j$ indicates the grid location, and $p_{occ}(i,j) = 1$ implies the grid cell is occupied. The global coordinates for a grid location $(i,j)$ are given by $( x(i), y(j) )$. Next, for each viewpoint in the cluster from VLPG, we use their camera fields-of-view to update their probability map $p_{nl}$. For each cell $(i,j)$ in the grid map, we update the probability as, 
\begin{multline}
p_{nl}(i,j) = p_{nl}(i,j) \cdot (1 - I(i, j, \mathbf{x}_{vp}) \cdot p_l^*) \\ \forall \; \mathbf{x}_{vp} \in C_{vp} \text{ and } p_{nl}(i,j) \neq 1,
\end{multline}
where
$$
  I(i,j, \mathbf{x}_{vp}) = {
  \begin{cases}
    1, & \text{for } (i,j) \in fov(\mathbf{x}_{vp}) \text{ and } p_{nl}(i,j) \neq 1 \\
    0, & else 
  \end{cases}}
$$
and $p_l^*$ is a user-defined probability threshold. After accounting for every viewpoint in the cluster, the probability of localizing the object at any grid cell can be computed as
$$
p_l(i,j) = 1 - p_{nl}(i,j).
$$

Consequently, the object localization probability map ($p_l$) provides a likelihood over the $6m \times 6m$ grid map for localizing the object of interest based on VLPG. From the probability map, we compute an alternative viewpoint to view the high-likelihood regions. To select an alternative viewpoint, we start by sampling a set of poses $\mathcal{S}_{vp}$ on the local map. A suitable alternate viewpoint is selected from the samples based on $p_l$ by minimizing against a cost function. We define the cost function as,
\begin{equation}
  J(\mathbf{x}_{vp}) = w_{d} \cdot J_{distance} + w_{q} \cdot J_{quality} + w_{obs} \cdot J_{obstacle}
\end{equation}
Here, $w_d, w_q, w_{obs}$ are user-defined weights. The distance cost is a measure of how far the new viewpoint is from the current position. The quality cost is a measure of how probable it is to view the object from the re-planned viewpoint. The collision cost encourages having clearance around the robot from the re-planned viewpoint.
\begin{align}
  J_{distance} = (\mathbf{x}_{vp} - \mathbf{x}_{r}) \; I \; (\mathbf{x}_{vp} - \mathbf{x}_{r})^T \\
  J_{quality} = \sum_{(x,y) \in fov(\mathbf{x}_{vp})} 1 - p_l(x,y) \quad \\
  J_{obstacle} = e^{-{d_o}^2} \quad\quad\quad\quad\quad\quad\quad\quad\quad
\end{align}
Here, $I$ is the identity matrix and $d_o$ is the distance to the closest obstacle from the replanned viewpoint.
\begin{align}
  \mathbf{x}_{vp} = \{\mathbf{x}_{vp} \mid J(\mathbf{x}_{vp}) = \min_{\mathbf{x}'_{vp} \in \mathcal{S}_{vp}} J(\mathbf{x}'_{vp})\}
\end{align}

The robot navigates to the chosen viewpoint. If the object is observed, the local planner corrects the robot's pose to keep the object centered (as discussed in Section 4A), and the task is considered a success. If the object is not in view at the replanned viewpoint, the probability values for the region currently in the camera view are reduced as
$$
p_{l}(x,y) = 0.2 \cdot p_l(x,y) \quad \forall (x,y) \in fov(\mathbf{x}_{vp}).
$$
The robot re-computes a new viewpoint using this updated probability map. The replanning process continues until the object is detected or maximum re-computations are reached. 


\begin{figure}[t]
\centering
\subfigure[VLPG Initial Viewpoint]{\label{fig:a}\includegraphics[width=0.40\linewidth, trim={0cm, 17cm, 0cm, 0cm}, clip]{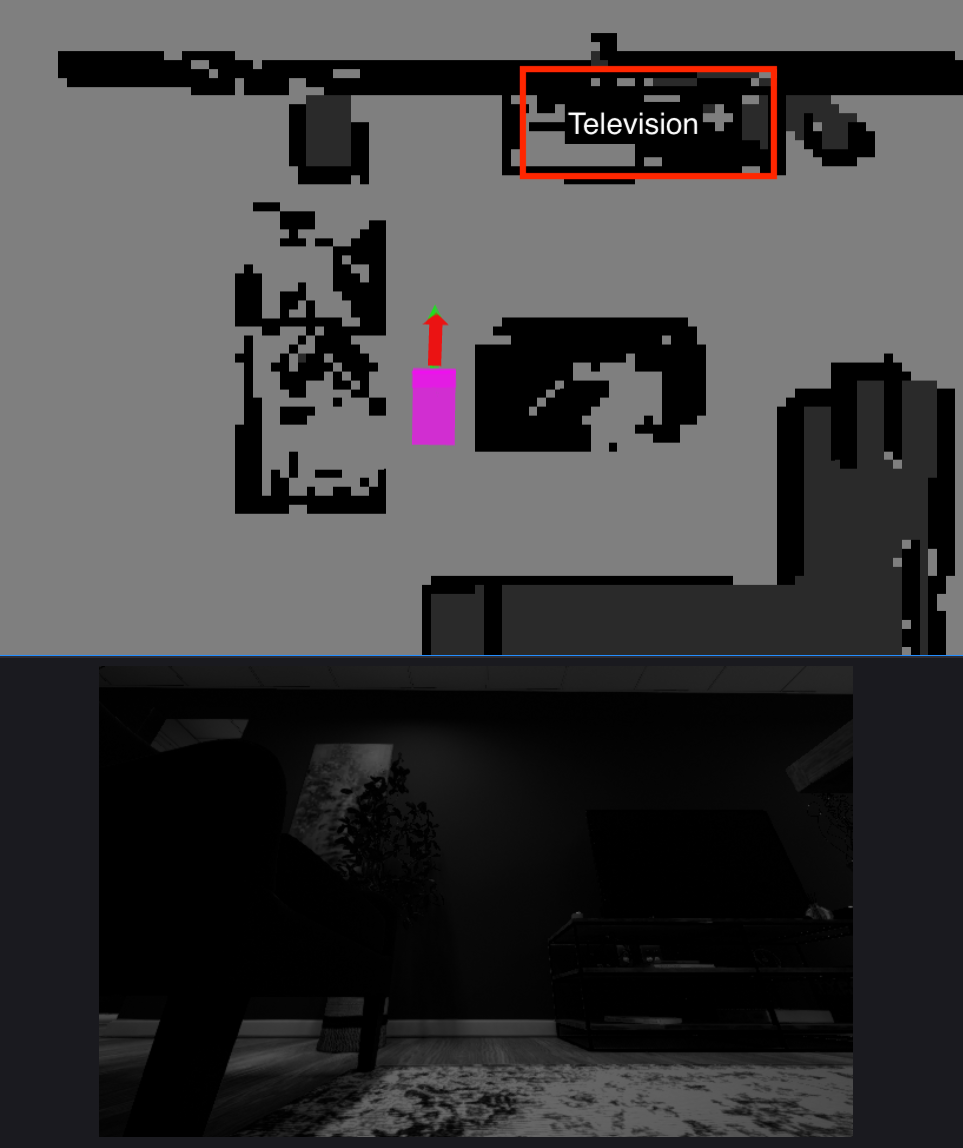}}
\subfigure[VLPG + Object Centering]{\label{fig:a}\includegraphics[width=0.40\linewidth, trim={0cm, 17cm, 0cm, 0cm}, clip]{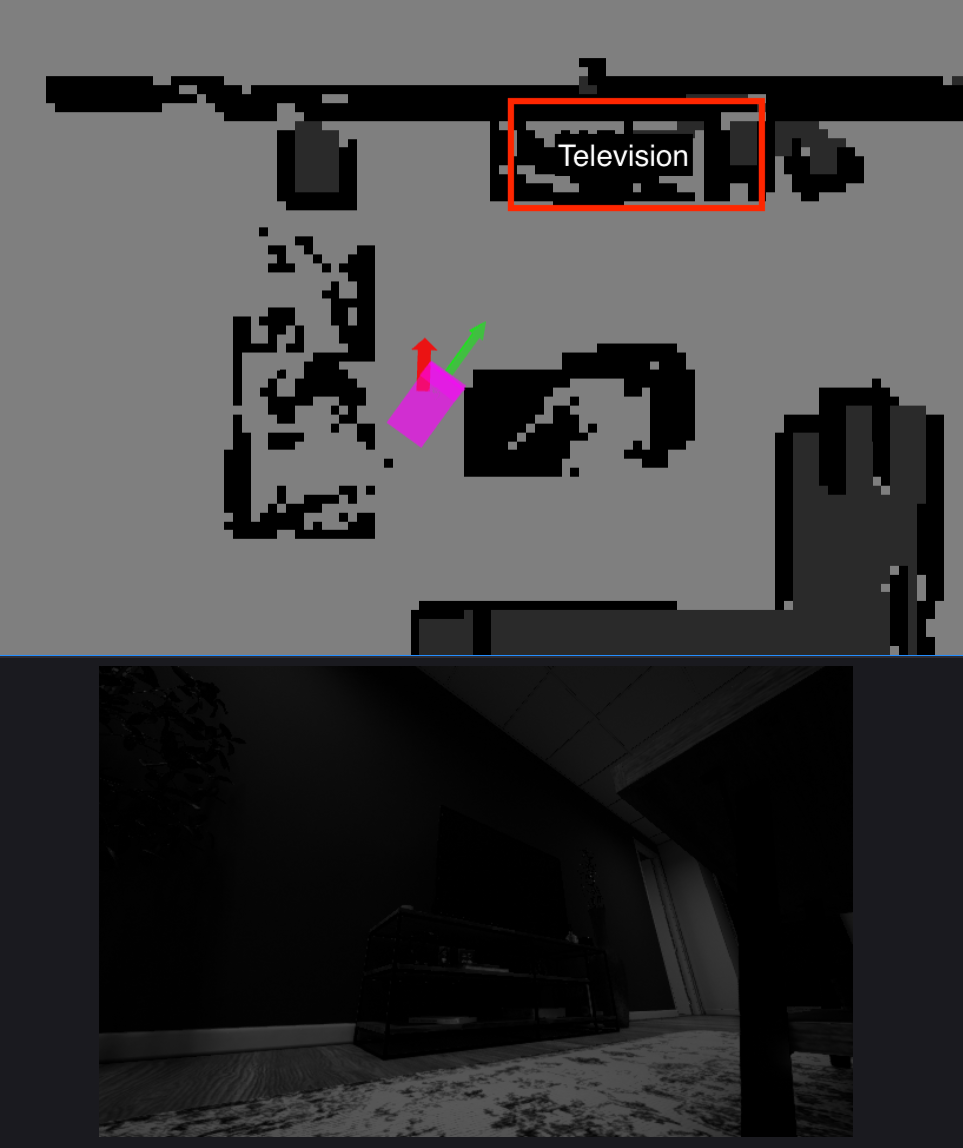}}
\caption{The red arrow shows the viewpoint computed as an initial guess, while the green arrow is the robot's pose at the end state. We can observe that the object centering directs the robot to the object of interest.}
\vspace{-5pt}
\end{figure}

\begin{figure}[t]
\centering     
\subfigure[Initial viewpoint]{\label{fig:a}\includegraphics[width=23mm, trim={0cm, 15cm, 10cm, 5cm}, clip]{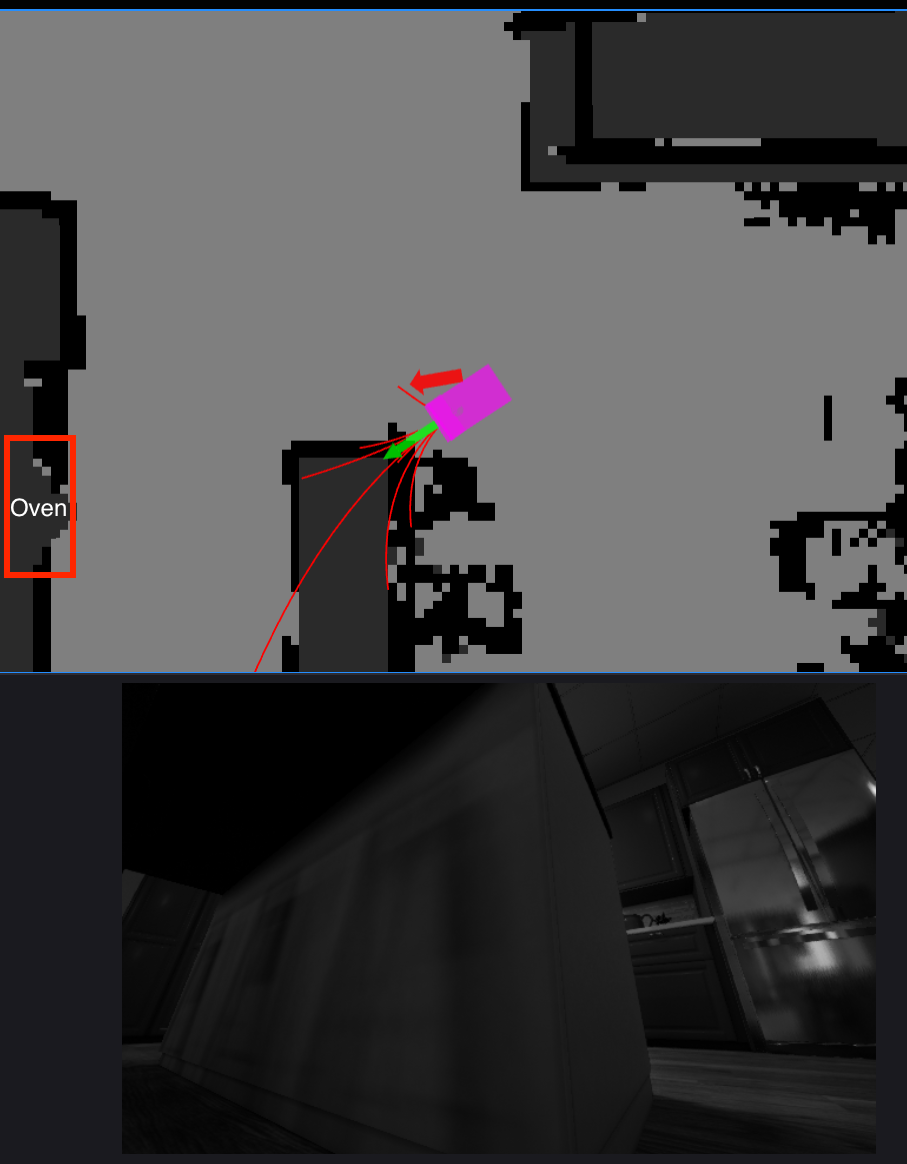}}
\subfigure[Re-planning]{\label{fig:a}\includegraphics[width=23mm, trim={0cm, 15cm, 10cm, 5cm}, clip]{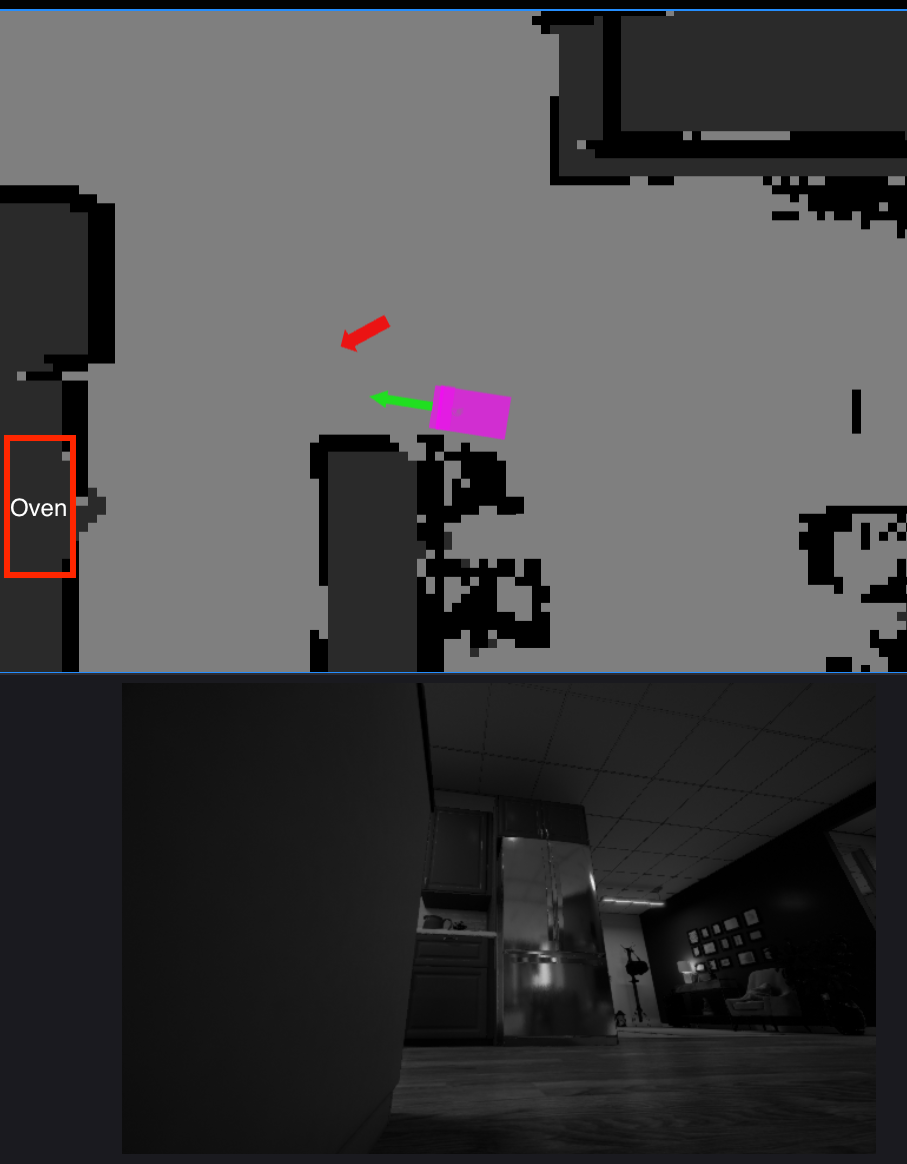}}
\subfigure[Final state]{\label{fig:a}\includegraphics[width=23mm, trim={0cm, 15cm, 10cm, 5cm}, clip]{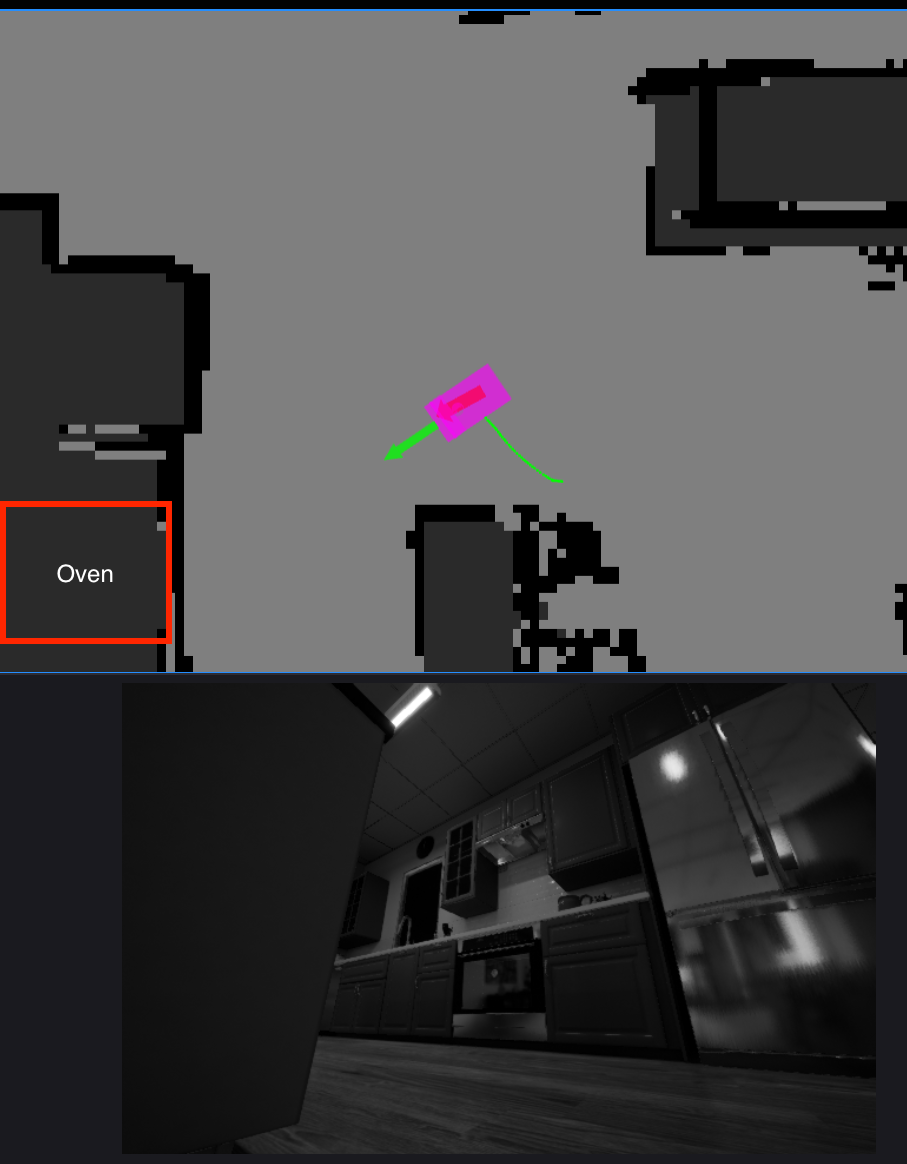}}\\
\caption{An example demonstrating ``Local Search", where the robot is assigned the task of viewing an {\em{oven}}. (Fig. 3-a) The initial viewpoint guess results in the robot view being occluded. (Fig. 3-b) The local search identifies a suitable alternative viewpoint (red), which gets a better view of the object. (Fig. 3-c) The robot then moves to the replanned viewpoint and successfully views the object.}
\vspace{-15pt}
\end{figure}

\begin{table}[t]
\resizebox{\linewidth}{!} {
\centering
 \begin{tabular}{|c|| c | c | c | c |} 
 \hline
 Object Prompts & Frontier Exploration & VLPG & VLPG + Obj. Centering & VLPG-Nav \\
  & (Baseline 1) & (Baseline 2) & (Baseline 3) & (Proposed)\\
 \hline\hline
 Bedside Lamp & - & 144 & 9 & 9 \\ 
 Cradle & - & 33 & 7 & 7 \\
 Television & - & 108 & 10 & 10 \\
 Beanbag & - & 96 & 8 & 8 \\
 Oven & - & - & - & 22 \\ 
 Buddha Statue & 15 & - & - & - \\ 
 Bar Stool & - & 12 & 3 & 3 \\ 
 Coffee Table & - & 31 & 4 & 4 \\  
 \hline
 \end{tabular}
}
 \caption{We tabulate the distance of the bounding box centers from the center of the image (in pixels). Since the bounding box information is noisy, we average the bounding box centers and round it to the nearest integer. Frontier exploration found the Buddha statue as it was located near the robot's start position. Though the proposed method was unable to view the Buddha statue in simulation (as the initial guess was incorrect), it was able to view it in our real-world experiments. }\label{tab:bb}
 \vspace{-5pt}
\end{table}
\begin{table}[t]
\resizebox{\linewidth}{!} {
\centering
\begin{tabular}{| c || c | c | c | c |} 
\hline
Object Prompts & Frontier Exploration & VLPG & VLPG + Obj. Centering & VLPG-Nav \\
& (Baseline 1) & (Baseline 2) & (Baseline 3) & (Proposed)\\
\hline\hline
Bedside Lamp & - & 43.86 & 5.12 & 5.12 \\ 
Cradle & - & 21.92 & 15.58 & 15.58 \\
Television & - & 50.66 & 1.77 & 1.77 \\
Beanbag & - & 24.14 & 9.34 & 9.34 \\
Oven & - & - & - & 22.2 \\ 
Dishwasher & - & - & - & -\\ 
Buddha Statue & 23.31 & - & - & - \\ 
Bar Stool & - & 0.66 & 4.47 & 4.47 \\ 
Coffee Table & - & 34.15 & 28.58 & 25.58 \\  
\hline \hline
SAE ($\uparrow$) & 0.12 & 0.66 & 0.73 & {\bf{0.85}} \\
\hline
\end{tabular}
}
\caption{We tabulate the angle made by the robot's orientation with the line of sight between the robot and the object's ground truth positions. This angular value denotes the error in centering the object in the camera field of view. The last row lists the success-weighted measure of angular error (SAE). We highlight the proposed method (VLPG-Nav) results in the best SAE. Frontier exploration was able to view the Buddha statue as it was partly in view from the robot's starting position. Though the proposed method failed to view the Buddha statue in simulation (as the initial guess was incorrect), it was viewed successfully in our real-world experiments.}\label{tab:gt}
\vspace{-15pt}
\end{table}

\begin{figure}[t]
\centering     
\subfigure[Floor Plan]{\label{fig:a}\includegraphics[height=2.8cm,width=0.14\textwidth]{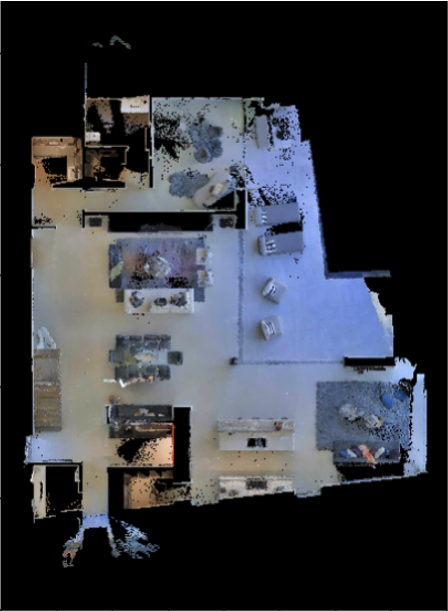}}
\subfigure[Ground Truth]{\label{fig:a}\includegraphics[height=2.8cm,width=0.14\textwidth]{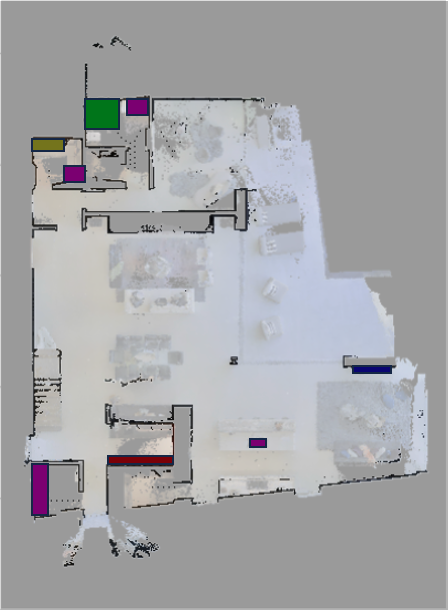}}
\subfigure[VLMap~\cite{vlmap}]{\label{fig:a}\includegraphics[height=2.8cm,width=0.14\textwidth]{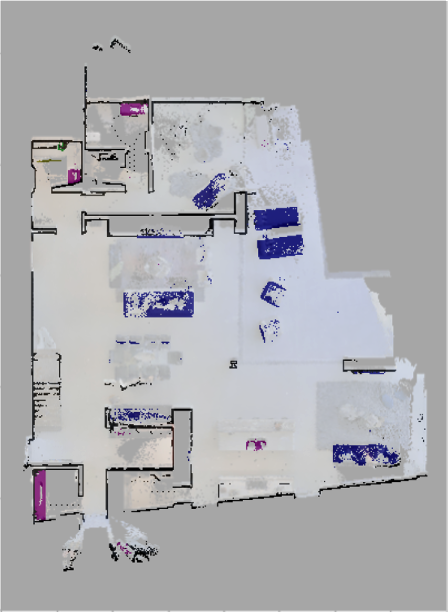}}
\subfigure[VLPG-Nav (ours)]{\label{fig:a}\includegraphics[height=2.8cm,width=0.14\textwidth]{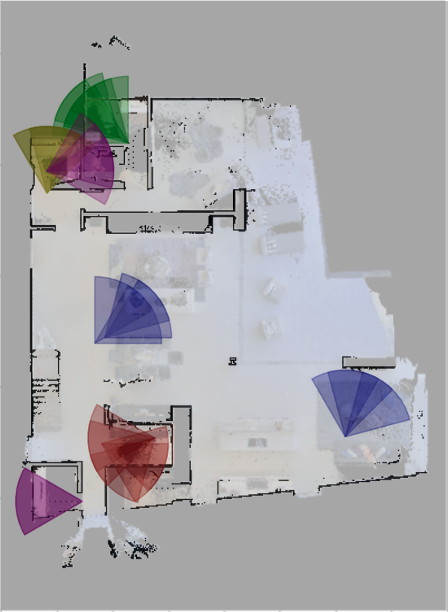}}
\subfigure[Legend]{\label{fig:a}\includegraphics[width=0.14\textwidth]{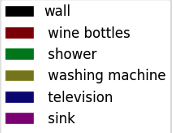}}
\caption{
An illustrative example of the viewpoints generation using prior environment knowledge. (Fig. 5-a) The environment's floor plan. (Fig. 5-b) Ground truth locations of the object of interest on the 2D floor plan. (Fig. 5-c) The 2D object localization computed using VLMap~\cite{vlmap}. (Fig. 5-d) The figure represents the object viewpoint and camera FoV generated by our approach. In this case, the walls (depicted in black) are from the occupancy map and are not computed using the VLPG. (Fig. 5-e) Depicts the objects of interest with specific colors used on the map. The proposed method improves object localization and consequently object goal navigation since object viewpoints are linked to the SLAM pose graph, which gets optimized over time. Computationally, VLPG is memory efficient and does not require depth maps for 2D projection, making it effective for deployment on low-compute robots.
}\label{fig:mapping}
\vspace{-5pt}
\end{figure}
\begin{figure}
\centering
\subfigure[Bed]
{\label{fig:a}\includegraphics[width=0.18\textwidth]{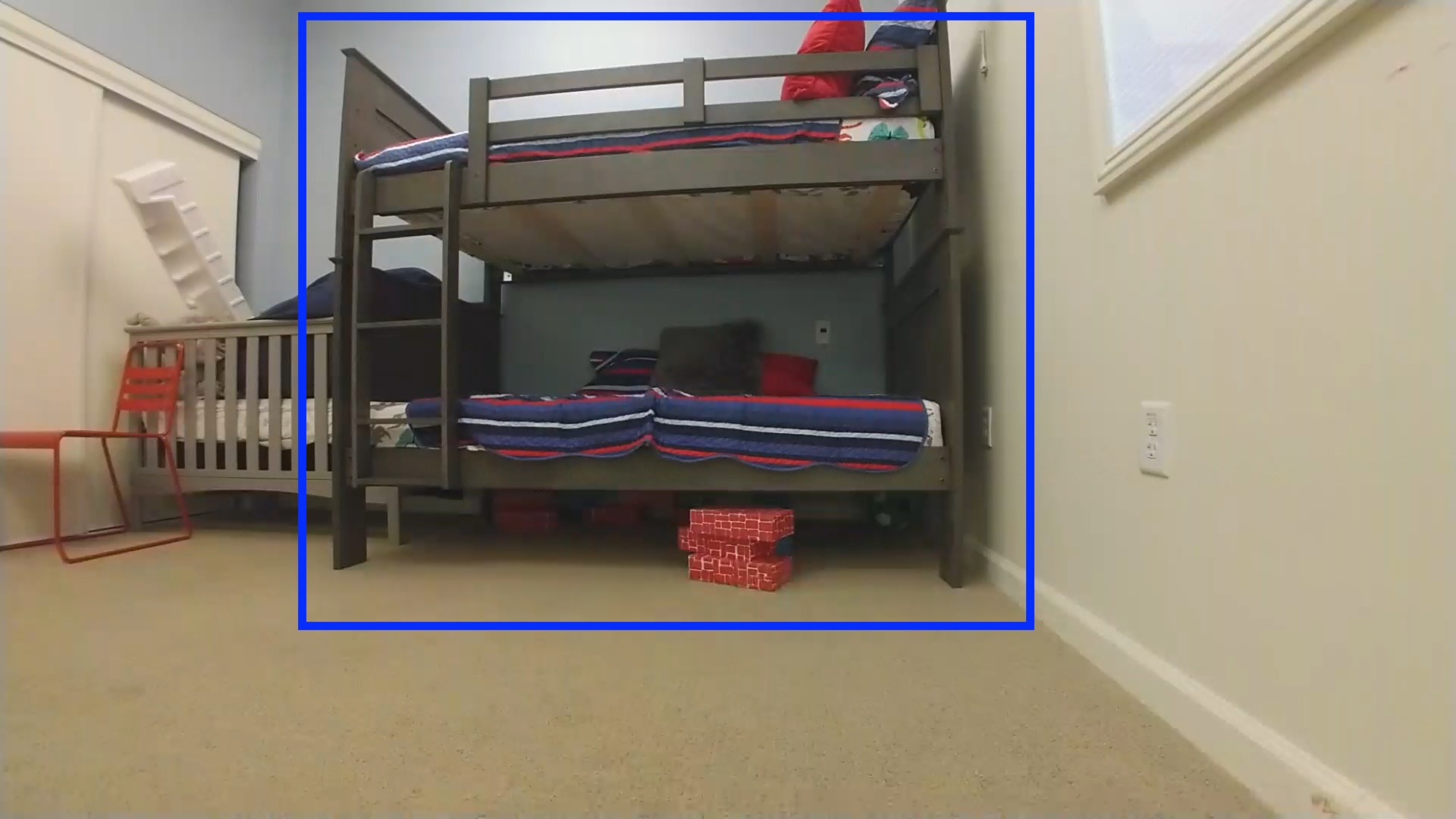}}
\subfigure[Buddha Statue]{\label{fig:a}\includegraphics[width=0.18\textwidth]{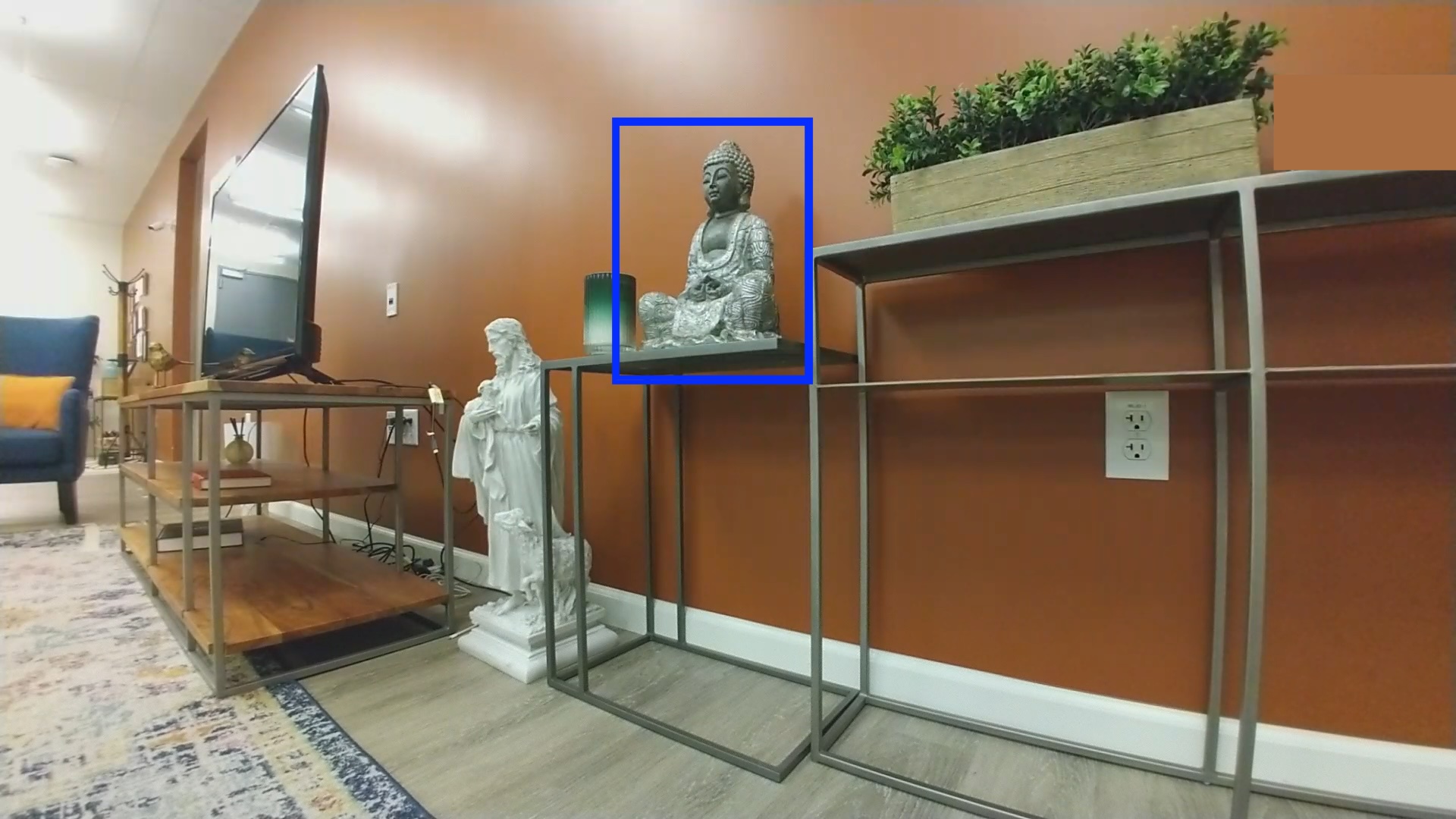}}
\subfigure[Television]{\label{fig:a}\includegraphics[width=0.18\textwidth]{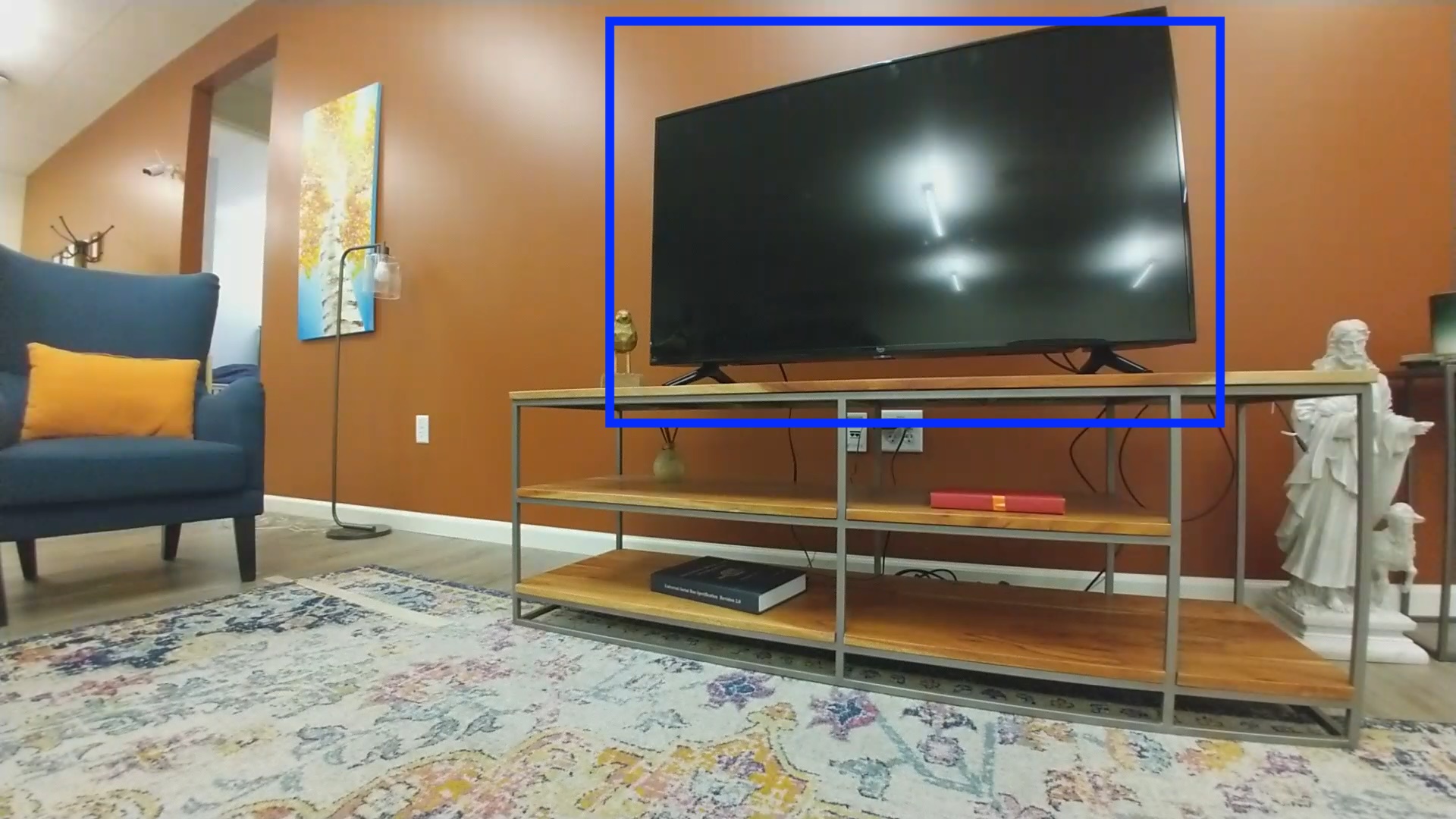}}
\subfigure[Plant]{\label{fig:a}\includegraphics[width=0.18\textwidth]{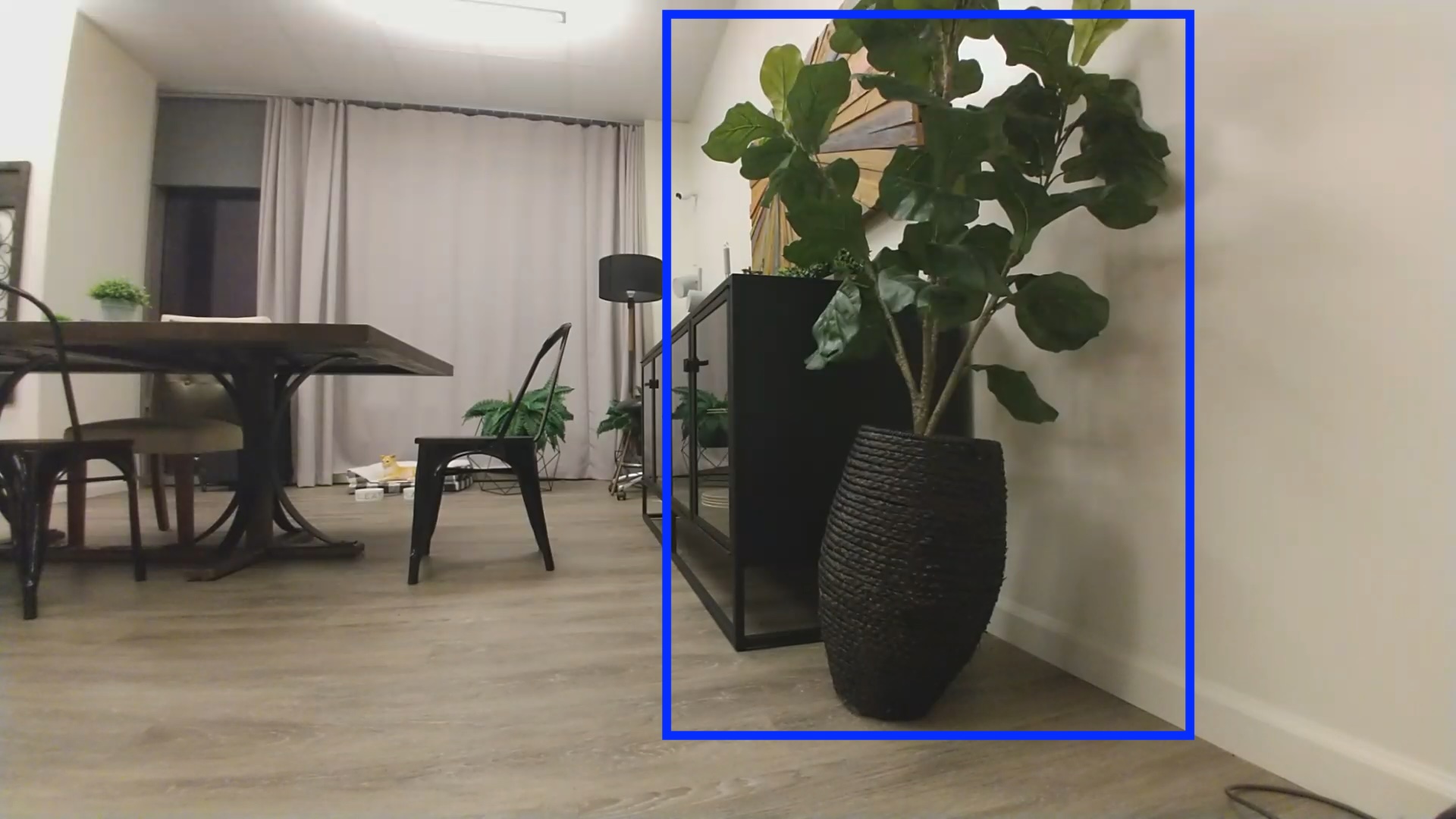}}
\caption{{\bf{Real-World Experiment:}} 
We include snapshots from the robot’s camera captured at the conclusion of the object navigation task. The robot successfully reached the considered objects (Bed, Buddha Statue, Television, Plant). To highlight the objects of interest, bounding boxes have been added to the figures.}\label{fig:realworld_success}
\vspace{-15pt}
\end{figure}
\section{Evaluations}
In this section, we outline our experimental setup and highlight the performance of our approach, VLPG-Nav. 


\subsection{Experimental Setup}
We evaluate our proposed method (Section V-C) in both simulation and the real world by considering a non-holonomic robot and different objects of interest. Our simulation uses Gazebo and Unreal Engine, where Unreal simulations provide photo-realistic images for the robot
, and an equivalent home environment in Gazebo is used for simulating the robots. Our object localization uses the ResNet50 version of CLIP and utilizes GradCAM to compute bounding boxes of regions of interest. The local planning optimization is implemented using the NLOPT library. For real-world experiments, our mobile ground robot used two Qualcomm QCS605 octa-core ARM processors, each with 2GB RAM and with two cores running at 2.5GHz and six cores running at 1.7GHz. The local search replanning, CLIP-based evaluations for computing the initial viewpoint guess, and object centering are performed on a laptop equipped with an Intel Core-i7 6th generation processor, utilizing sensor inputs from the robot. The computed object bounding boxes and viewpoints are then communicated back to the robot which execute object centering and navigation.

\subsection{Evaluation Metrics}
Prior object navigation works~\cite{vlmap} have typically defined success as the robot reaching within a specified distance threshold of its target object. However, our approach considers success when the robot successfully navigates to the object of interest and centers it within its camera view. This accounts for occlusion and ensures that success is only acknowledged when the object is visible to the robot's camera.
In this regard, we define a new evaluation metric: the Success weighted measure of Angular Error (SAE). This metric quantifies the angular deviation between the robot's orientation and the line of sight between the robot and the object's center.
\vspace{-10pt}
\begin{equation}
    SAE = \frac{1}{N} \sum^N S_i \cdot \exp^{- (\frac{\Delta \theta} { \Delta \theta_{max}})^2}
\end{equation}
Here, $S_i$ is an indicator function that has a value of $1$ when the object is in the camera field of view, else is $0$. $\Delta \theta_{max}$ is set to $90\deg$ and is a measure of the horizontal field of view of the camera. An SAE of $1$ indicates that the robot viewed and centered the object in its camera view in every instance. While an SAE of $0$ indicates the robot never succeeds in viewing the object.  

\subsection{Object Navigation}
We consider a household environment where the robot is given the task of observing objects of interest. In our simulation environment, we consider eight randomly selected objects of interest and evaluate our proposed VLPG-Nav method against three baseline approaches. The first baseline adopts a purely exploratory method, utilizing frontier exploration~\cite{frontier} and CLIP~\cite{clip}+GradCAM~\cite{gradcam} to identify and approach the object. The second baseline uses the VLPG to select an initial viewpoint and guide the robot towards it. The third baseline uses VLPG for initial viewpoint selection and combines with our object-centering method to refine the final pose for optimal object viewing. The proposed approach (VLPG-Nav) uses VLPG and object centering and performs a local search to find the object when the object is not in view. The evaluation serves as an ablation study highlighting the importance of the three components of our proposed method for reliable object goal navigation. We regard the execution as a failure when an incorrect object is detected or the object is not detected within a maximum simulation time of 5 minutes.
 
In Table~\ref{tab:bb}, we compare the methods in terms of their ability to view the object head-on. Since our method uses the bounding box information to center the object, we tabulate the error in terms of the pixel distance between the bounding box centroid and the image center along the horizontal image axis. We observe that the object-centering component results in a smaller error than relying purely on the VLPG initial viewpoint guess. 

Since the bounding boxes are generated using GradCAM, they need not cover the entire object of interest. In Table~\ref{tab:gt}, we use the ground truth positions of the robot and the object from the end state to measure the angular deviation between the robot's line of sight and the object center. On average, we observe that the object-centering cost function helps to center the object in the image. Moreover, we tabulate the SAE values in the table. We observe the proposed method results in the best SAE values compared to the baselines.

In the case of an {\em{oven}}, the kitchen counter obstructs the robot's view of the oven at the end state. This results in failure to view the object in all baseline methods except for the proposed VLPG-Nav. The local search helps identify an alternative viewpoint to bring the oven into view. In the case of the Buddha statue, the initial viewpoint guess was in a completely different area of the environment, and local search could not help bring the object into view. As can be observed in Fig.~\ref{fig:realworld_success}-b, the robot navigated successfully to the Buddha statue in our real-world experiment.

In the case of random exploration, many false positives resulted in incorrectly identifying the object. In the case of the Buddha statue, random exploration was successful as the statue was near the robot's starting position.

\subsection{Object Localization}
In Fig.~\ref{fig:mapping}, we illustrate the viewpoints clusters computed by VLPG and cluster (Fig.~\ref{fig:mapping}-d) for a set of objects in a sample environment from Matterport 3D dataset~\cite{Matterport3D}. We also show the ground truth and VLMap~\cite{vlmap} generated maps as comparisons. In VLMaps, the objects are localized in the 2D map, while in our case, we do not localize the object exactly in the 2D space but instead find the viewpoint for the object in the 2D space which provides an approximation of object locations in the environment, and is used in our probability map construction for local search. 

\subsection{Real-world Experiments}
We evaluated our approach in real-world scenarios by considering a set of object prompts. Fig.~\ref{fig:realworld_success} illustrates the robot's camera view after navigating and framing the different objects. We observe that for the prompts {\em{bed, Buddha statue}} the objects are centered in the camera view. While for {\em{Television, Plant}} the object has a small angular deviation between the robot's orientation and the object's line of sight. In Fig. 1(b-d), we showcase the behavior when the chosen initial object viewpoint is occluded. We occlude the object {\em{plant}}, and our method successfully replans an alternative viewpoint using the object localization probability map.


\subsection{Discussion}

We highlight that our method is not specific to CLIP and can be replaced with other models such as GLIP~\cite{glip}, FLAVA~\cite{flava}, etc. In our implementation, we utilize CLIP for its comparatively faster online computation as we require the bounding box computation in our object-centering component to run at approximately 10Hz for our local planner.

\section{CONCLUSIONS}
We presented VLPG-Nav, a visual language navigation framework for object navigation. We build a visual language pose graph (VLPG) as a sparse mapping of the object viewpoints in the environment and is used to guess a relevant viewpoint given an open-vocabulary object prompt. Our proposed pose correction methods successfully replan to avoid object occlusions that prevent object detections and orient the robot to center the object in the camera view. 
Our method has some limitations. First, the success of our local search phase depends on the quality of the initial {\em{top-k}} viewpoints. 
Second, CLIP + GradCAM can be noisy for object detection and result in incorrect attention maps in the image. Moreover, the cost function for selecting an alternative viewpoint during the local search does not model the non-holonomic dynamics of the robot. In future work, we plan to explore better cost functions to choose alternative viewpoints in the probability map. Specifically, we plan to formulate it as a function of the local planner's trajectory cost~\cite{park_mpepc} to incorporate the kinematic limits during viewpoint re-planning. Moreover, we plan to explore alternative embedding compared to CLIP for constructing the VLPG.
\bibliographystyle{IEEEtran}
\bibliography{IEEEabrv,references}

\end{document}